\documentclass{article}
\usepackage{ijcai25}

\usepackage{times}
\usepackage{soul}
\usepackage{url}
\usepackage[hidelinks]{hyperref}
\usepackage[utf8]{inputenc}
\usepackage[small]{caption}
\usepackage{graphicx}
\usepackage{amsmath}
\usepackage{amsthm}
\usepackage{booktabs}
\usepackage{algorithm}
\usepackage{algorithmic}
\usepackage[switch]{lineno}
\usepackage{amsfonts}
\usepackage{bbm}

\usepackage{xcolor}

\usepackage{pifont}  
\usepackage{adjustbox}
\usepackage{booktabs} 
\usepackage{multirow}

\newcommand{\cmark}{\ding{51}} 

\title{A Comprehensive Survey on Concept Erasure in \\ Text-to-Image Diffusion Models}

\author{
Changhoon Kim\textsuperscript{1}
\And
Yanjun Qi\textsuperscript{2}
\\
\affiliations
\textsuperscript{1}Soongsil University, \textsuperscript{2}University of Virginia\\
\emails
kch@ssu.ac.kr, yanjun.research@gmail.com
}

\begin{document}

\maketitle

\begin{abstract}
Text-to-Image (T2I) models have made remarkable progress in generating high-quality, diverse visual content from natural language prompts. However, their ability to reproduce copyrighted styles, sensitive imagery, and harmful content raises significant ethical and legal concerns. Concept erasure offers a proactive alternative to external filtering by modifying T2I models to prevent the generation of undesired content. In this survey, we provide a structured overview of concept erasure, categorizing existing methods based on their optimization strategies and the architectural components they modify. We categorize concept erasure methods into fine-tuning for parameter updates, closed-form solutions for efficient edits, and inference-time interventions for content restriction without weight modification. Additionally, we explore adversarial attacks that bypass erasure techniques and discuss emerging defenses. To support further research, we consolidate key datasets, evaluation metrics, and benchmarks for assessing erasure effectiveness and model robustness. This survey serves as a comprehensive resource, offering insights into the evolving landscape of concept erasure, its challenges, and future directions.

\end{abstract}

\begin{figure}[t]
    \centering
    \includegraphics[width=\linewidth]{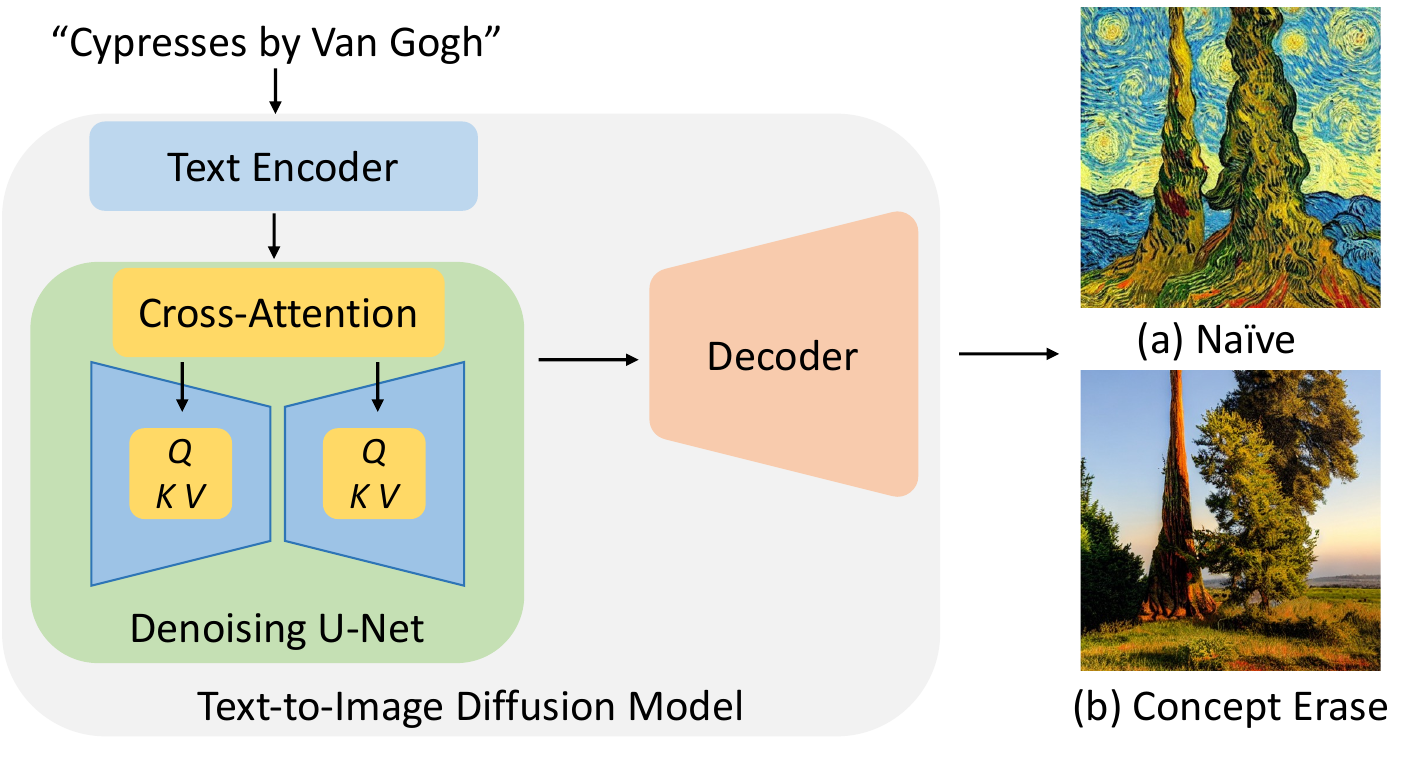}
    \caption{ Concept Erasure for Text-to-Image Diffusion Model Pipeline. The model synthesizes images from textual prompts. In Figure (a), the model faithfully follows the input prompt, generating an image that aligns with the specified concept. In contrast, Figure (b) illustrates the effect of concept erasure, where the model suppresses the erased concept, ``Van Gogh’’, ensuring its exclusion from the generated output. For a detailed explanation, refer to Sec.~\ref{sec:preliminaries}.
    }
    \label{fig:overview}
\end{figure}

\section{Introduction}

Recent advancements in Text-to-Image (T2I) models significantly enhance their ability to generate high-quality images that align closely with user-provided textual prompts~\cite{stable_diffusion}. These models empower users with unprecedented creative freedom, enabling the production of sophisticated and photorealistic images without requiring professional expertise. However, as T2I models are trained on vast text-image datasets scraped from the internet, they inherit both the strengths and biases of their training data. Specifically, they can replicate copyrighted artistic styles, explicit content, and private or sensitive visual concepts~\cite{Somepalli2023UnderstandingAM,Somepalli2022DiffusionAO},
raising ethical and legal concerns. The unconstrained generative capabilities of these models can be exploited for malicious purposes, such as creating misleading or harmful content, generating deepfake imagery, or manipulating public opinion.

To address these concerns, researchers have developed post-hoc safety mechanisms that integrate security protocols into the T2I model pipeline~\cite{WOUAF,stable_signature,kim2020decentralized,nie2023attributing,tree_ring}. These methods include watermarking, model attribution, and forensic tracking techniques that help identify the sources responsible for AI-generated content. While these approaches provide valuable forensic tools for mitigating misuse, they remain \textit{reactive solutions} that only take effect after potentially harmful images have been generated and disseminated. 


Given the constraints of reactive methods, an emerging line of research explores \textit{proactive solutions}, specifically concept erasure, which aims to systematically remove targeted concepts from a model’s generative capability. As illustrated in Fig.~\ref{fig:overview}, concept erasure techniques suppress a model’s ability to generate protected or undesired content, ensuring that even when explicitly prompted, the model does not produce outputs containing erased concepts. These methods operate by either modifying the model’s internal components or intervening in the inference process, preventing the unauthorized reproduction of sensitive or restricted concepts. 
Other related approaches include image editing, which modifies attributes within a given image. In contrast, concept erasure enforces persistent modifications that prevent the generation of targeted concepts across all inputs~\cite{Arad2023ReFACTUT}.

Research on securing T2I models has gained significant attention, leading to comprehensive surveys on various aspects of generative model security. 
Recent studies from \cite{Zhang2024AdversarialAA} and \cite{Truong2024AttacksAD} focus on surveying adversarial attacks and defenses, with the former specifically addressing T2I diffusion models and the latter examining threats in diffusion-based image generation. While these works offer valuable insights, they do not provide a focused analysis of concept erasure in T2I models. The absence of a dedicated survey makes it difficult to systematically understand and compare existing concept erasure techniques. 

\begin{figure}[t]
    \centering
    \includegraphics[width=\linewidth]{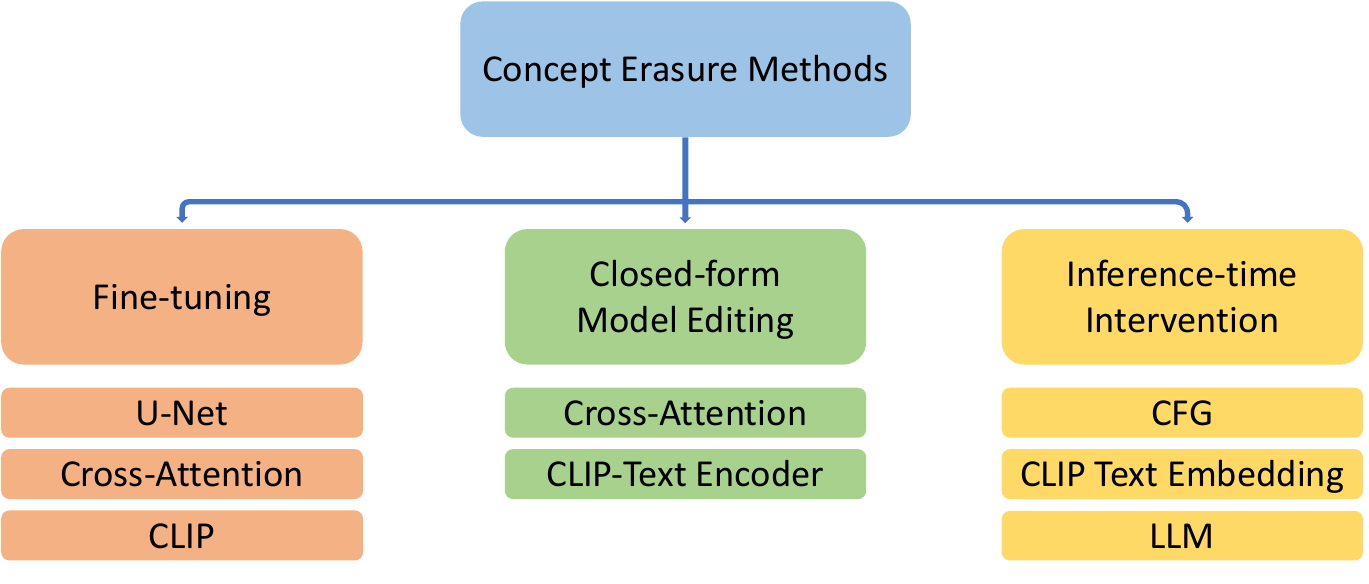}
    \caption{
    Taxonomy of Concept Erasers. Concept erasure methods are categorized based on their optimization strategy (first level) and the model components they modify (second level). A detailed discussion is provided in Sec.~\ref{sec:method}.
    }
    \label{fig:taxonomy}
\end{figure}

To bridge this gap, we present a structured and comprehensive survey on concept erasure techniques in T2I diffusion models, offering the following key contributions:
\begin{itemize}
    \item \textbf{Comprehensive Taxonomy of Concept Erasure Methods.} We systematically classify existing concept erasure techniques based on their optimization strategies and the model components they modify, as illustrated in Fig.~\ref{fig:taxonomy}. This categorization provides a structured taxonomy for understanding different concept erasure approaches.
    
    \item \textbf{Analysis of Adversarial Attacks and Defense Mechanisms.} We investigate adversarial attacks designed to circumvent concept erasure, categorizing them based on access to the T2I model’s internal components. Additionally, we explore emerging defense strategies aimed at enhancing the robustness of concept erasure techniques.
    
    \item \textbf{Evaluation Benchmark for Concept Erasure.} We consolidate widely used datasets and metrics for assessing erasure effectiveness, model fidelity, and resilience against adversarial attacks, providing a standardized benchmark for future research and evaluation.
    
    \item \textbf{Future Research Directions and Open Challenges.} We introduce research opportunities in concept erasure, including improving generalization across different modalities such as text-to-video, developing fine-grained benchmarks to evaluate erasure effectiveness while capturing unintended distortions in related attributes, and addressing novel adversarial threats.
    
\end{itemize}

To provide a comprehensive understanding of concept erasure in T2I models, we structure this paper as follows. Sec.~\ref{sec:preliminaries} provides background knowledge on the architecture of T2I models and introduces key technical concepts relevant to concept erasure. Sec.~\ref{sec:method} categorizes concept erasure techniques based on their optimization methods and the model components they modify. Sec.~\ref{sec:robustness} examines adversarial attacks that attempt to circumvent concept erasure, along with emerging defense strategies designed to enhance robustness. Sec.~\ref{sec:evaluation} reviews commonly used evaluation metrics and datasets for benchmarking concept erasure methods. Based on these findings, Sec.~\ref{sec:future_research} discusses future research directions, open challenges, and opportunities in this field, followed by conclusions in Sec.~\ref{sec:conclusion}.

\section{Backgrounds} \label{sec:preliminaries}

This section presents an overview of the Text-to-Image (T2I) diffusion model with a particular focus on Stable Diffusion (SD)~\cite{stable_diffusion}. 
As shown in Fig.~\ref{fig:overview}, SD comprises three main components: a vision decoder for reconstructing images from latent representations, a latent diffusion model for iterative denoising, and a conditional text encoder that transforms textual prompts into conditioning vectors.
We outline both the training and inference mechanisms of SD, which are essential for understanding how concept erasure techniques modify key model components or inference steps to suppress undesired concepts.

\begin{table*}[t!]
  \caption{Taxonomy of Concept Erasure Methods in T2I Models. Methods are categorized based on their optimization strategies and the specific model components they modify. In the third column, "CA" denotes the cross-attention layers within the latent diffusion model, while "CFG" refers to Classifier-Free Guidance adjustments. A comprehensive discussion of these methods is provided in Sec.~\ref{sec:method}.}
  \centering
  \scriptsize
  \setlength{\tabcolsep}{3pt} 
  \begin{adjustbox}{width=\textwidth,center}
    \begin{tabular}{p{1.8cm}p{3.6cm}cccccp{6.5cm}} 
      \toprule
      \textbf{Category} & \textbf{Representative Works} & \multicolumn{5}{c}{\textbf{Optimization Space}} & \hspace{5mm} \textbf{Description} \\ 
      \cmidrule(lr){3-7} 
       & & \textbf{U-Net} & \textbf{CA} & \textbf{CLIP} & \textbf{LLM} & \textbf{CFG} & \\ 
      \midrule
      \multirow{13}{1.8cm}{\textbf{Fine-Tuning}}  
      & FMN~\cite{Zhang2023ForgetMeNotLT} &  & \cmark &  &  &  & Minimize attention activation to erase concepts. \\ 
      & AC~\cite{Ablating_Concept} &  & \cmark &  &  &  & Remaps erased concepts to general concepts. \\ 
      & SALUN~\cite{fan2024salun} & \cmark &  &  &  &  & Modifies influential weights to remove concepts. \\ 
      & ESD~\cite{esd} & \cmark & \cmark &  &  &  & Edits noise prediction to remove concepts. \\ 
      & DT~\cite{Ni2023DegenerationTuningUS} & \cmark &  &  &  &  & Degrades model’s ability to reconstruct erased concepts. \\ 
      & Geom-Erasing~\cite{Liu2023ImplicitCR} & \cmark &  &  &  &  & Uses geometric constraints for concept removal. \\ 
      & SA~\cite{Heng2023SelectiveAA} & \cmark &  &  &  &  & Continual learning-based forgetting approach. \\ 
      & IMMA~\cite{Zheng2023IMMAIT} & \cmark &  &  &  &  & Enhances robustness against unauthorized fine-tuning. \\ 
      & SAFE-CLIP~\cite{safe_clip} &  &  & \cmark &  &  & Fine-tunes CLIP with safe and unsafe text-image quadruplets. \\ 
      & Latent Guard~\cite{Liu2024LatentGA} &  &  & \cmark &  &  & Fine-tunes the CLIP text encoder with safe and unsafe pairs. \\ 
      & AdvUnlearn~\cite{Zhang2024DefensiveUW} &  &  & \cmark &  &  & Adversarial finetuning for CLIP text encoder. \\
      & Receler~\cite{Huang2023RecelerRC} &  & \cmark &  &  &  & Uses adapters to enhance robustness. \\
      & R.A.C.E~\cite{RACE} & \cmark  & \cmark &  &  &  & Adversarially fine-tunes U-Net for resilience. \\
      \midrule
      \multirow{7}{1.8cm}{\textbf{Closed-form Model Editing}}  
      & ReFACT~\cite{Arad2023ReFACTUT} &  &  & \cmark &  &  & Updates CLIP’s memory via low-rank edits. \\ 
      & TIME~\cite{Orgad2023EditingIA} &  & \cmark &  &  &  & Modifies CA projection matrices for concept editing. \\ 
      & UCE~\cite{Gandikota2023UnifiedCE} &  & \cmark &  &  &  & Simultaneously erases multiple concepts. \\ 
      & MACE~\cite{Lu2024MACEMC} &  & \cmark &  &  &  & Utilize adapters for large-scale concept erasure. \\ 
      & EMCID~\cite{Xiong2024EditingMC} &  &  & \cmark  &  &  & Large-scale concept erasure via two-stage closed-form editing \\ 
      & MUNBa~\cite{Wu2024MUNBaMU} &  &  & \cmark &  &  & Uses Nash bargaining for controlled concept removal. \\ 
      & RECE~\cite{Gong2024ReliableAE} &  & \cmark &  &  &  & Integrates adversarial fine-tuning with closed-form editing. \\
      \midrule
      \multirow{6}{1.8cm}{\textbf{Inference-Time Intervention}}  
      & SLD~\cite{sld} &  &  &  &  & \cmark & Incorporates safety guidance to mitigate undesired concepts. \\ 
      & AMG~\cite{Chen2024TowardsMD} &  & &  &  & \cmark & Introduces three guidance strategies to prevent memorization. \\ 
      & SAFREE~\cite{safree} &  &  & \cmark &  &  & Self-validating filtering and re-attention for safe generation.\\ 
      & Content Suppression~\cite{Li2024GetWY} &  &  & \cmark &  &  & Adjusts embeddings to suppress concept generation. \\ 
      & ORES~\cite{ores} &  &  &  & \cmark & \cmark & Utilizes LLMs to filter and rewrite prompts for safer generation. \\ 
      & GuardT2I~\cite{Yang2024GuardT2IDT} &  &  &  & \cmark &  & Propose conditional LLM to detect adversarial prompts. \\ 
      & DNG~\cite{koulischer2024dynamic} &  &  &  &  & \cmark  & Introduce dynamic guidance to suppress undesired concepts.\\ 
      & TFSD~\cite{Kim2025TrainingFreeSD} &  &  & &  & \cmark & Propose safe denoiser to guide sampling away from unsafe regions.\\ 
      \bottomrule
    \end{tabular}
  \end{adjustbox}
  \label{tab:taxonomy}
\end{table*}


\subsection{Three Components of Stable Diffusion }

Stable Diffusion comprises three primary components:

\paragraph{(1) Image Autoencoder.} The model leverages a pre-trained autoencoder to compress high-dimensional image data into a low-dimensional latent representation. The encoding network $\mathcal{E}(\cdot)$ maps an image $x$ to a latent variable $z = \mathcal{E}(x)$, and the decoding network $\mathcal{D}(\cdot)$ reconstructs the image from the latent space such that $\mathcal{D}(z) = \hat{x} \approx x$. This design ensures effective data compression while minimizing reconstruction error, preserving essential image features critical for generative tasks.

\paragraph{(2) Latent Diffusion Model.} The core generative process in SD is governed by a U-Net-based Latent Diffusion Model (LDM) that progressively refines noisy latent representations toward high-fidelity outputs. The training objective is formulated as:
\begin{equation}
    L_{\text{SD}} = \mathbb{E}_{n \sim \mathcal{N}(0,1), z, c, t} \left[
    \| n - \Phi_{\theta}(z_t, c) \|_2^2
    \right],
\end{equation}
where $c$ is the text embedding derived from the input prompt and integrated via cross-attention, $t$ denotes the diffusion timestep, $n$ is a noise vector sampled from a standard Gaussian distribution $\mathcal{N}(0,1)$, and $z_t$ is the noisy latent variable at timestep $t$. The LDM \( \Phi_{\theta} \), parameterized by \( \theta \), is trained to predict and remove noise at each step, progressively refining the latent variable along the diffusion trajectory.

\paragraph{(3) Conditional Text Encoding.} The model employs a text encoder to transform user-provided text prompts into conditioning vectors, enabling fine-grained control over the generation process. Specifically, the textual prompt $y$ is embedded as $c = \mathcal{E}_{\text{txt}}(y)$, where $\mathcal{E}_{\text{txt}}$ typically textual encoder of CLIP~\cite{CLIP}. These text embeddings are integrated through the cross-attention layers within the latent diffusion model~\cite{stable_diffusion}, allowing the textual context to dynamically influence each denoising step.

\subsection{Inference in Stable Diffusion}
Classifier-free guidance~\cite{ho2022classifier} enhances the conditionality of the image synthesis process during the inference phase of SD. The process starts with initializing latent representations $z_T$ sampled from a Gaussian distribution. The denoising trajectory is steered by classifier-free guidance, which modifies the denoising function as follows:
\begin{equation}\label{eq:classifier-free-guidance}
    \Tilde{\Phi}_{\theta}(z_t, c) = \Phi_{\theta}(z_t, \phi) + \alpha \left( \Phi_{\theta}(z_t, c) - \Phi_{\theta}(z_t, \phi) \right),
\end{equation}
where $\Phi_{\theta}(z_t, c)$ and $\Phi_{\theta}(z_t, \phi)$ represent the conditioned and unconditioned latent noises, respectively. The guidance scale $\alpha > 1$ amplifies the influence of the conditioned path, embedding the textual information into the generative process. 
Iterative refinement reduces noise through sequential calculations of $z_{t-1} = \Tilde{\Phi}_{\theta}(z_t, c)$, progressing until $t=0$. The final coherent image representation $z_0$ is transformed into the output image $\hat{x}$ by the decoder, $\hat{x} = \mathcal{D}(z_0)$. The T2I generation process can be succinctly expressed as $SD(y) = \mathcal{D}(\Tilde{\Phi}_{\theta}(z_T, \mathcal{E}_{\text{txt}}(y)))$.

\section{Methods} \label{sec:method}

\subsection{Concept Erase}
Concept erasure in T2I models, particularly SD, involves modifying model parameters or adjusting inference procedures to selectively suppress or eliminate the generation of specific, unwanted concepts. This technique is crucial for addressing the risks associated with generating potentially harmful or copyrighted content in the model's outputs. The primary goal of concept erasure is to condition the model so that it does not produce images corresponding to undesired prompts. For instance, to erase the influence of a copyrighted artist's style, the model is adjusted such that a prompt like ``A painting in the style of [artist]'' results in outputs that bear no resemblance to that artist's work. This objective can be succinctly stated as \(SD(y_{\text{erase}}) \not\in  \{ x_{\text{erase}}\}\), where \(y_{\text{erase}}\) is the prompt that includes the concept to be erased, and \(x_{\text{erase}}\) denotes any image typically representative of that concept.

Concept erasure can be achieved through various optimization methods. And in each optimization group, methods can get classified by which components are modified to achieve the goal. This section,  therefore, categorizes existing methods by their optimization strategies and components they modify.
A comprehensive taxonomy with detailed explanations is provided in Tab.~\ref{tab:taxonomy}.
For a detailed explanation of each component and inference stage, please refer to Sec.~\ref{sec:preliminaries}.

\subsection{Fine-tuning Methods} \label{subsec:finetuning}
Fine-tuning is one of the most intuitive methods to erase undesired concepts from the T2I models. These methods iteratively optimize weights of component of Stable Diffusion (SD) to match erasing concept to its designed corresponding concept. For example, match representation of erasing concept $c_{erase}$, ``Van Gogh'' to $c_{target}$, ``Artist''. We categorize this by which component is updated to erase concept.

\paragraph{Fine-tuning  Latent Diffusion Model.}  To edit SD's Latent Diffusion Models (LDM) component, 
fine-tuning-based concept erasure methods selectively update model parameters to remove undesired concepts while preserving overall generative capabilities. These approaches can get categorized further based on the specific LDM's model components they modify, as different architectural elements govern distinct aspects of the image synthesis process.

A general formulation of fine-tuning for concept erasure in LDMs is as follows: \begin{equation}
    \min_{\theta} \left\| \Phi_{\theta}(z_t, c_{\text{erase}}) - \Phi_{0}(z_t, c_{\text{target}}) \right\|^{2}_2,
\end{equation}
where \( \Phi_{0} \) represents the pretrained LDM model, and \( \Phi_{\theta} \) denotes the fine-tuned LDM with updated parameters \( \theta \). The terms \( c_{\text{erase}} \) and \( c_{\text{target}} \) correspond to the text embeddings \( \mathcal{E}_{\text{txt}}(y_{\text{erase}}) \) and \( \mathcal{E}_{\text{txt}}(y_{\text{target}}) \), respectively. The objective enforces alignment between the erased concept \( c_{\text{erase}} \) and the target concept \( c_{\text{target}} \), ensuring that the model learns to replace undesired representations in the latent space.

One class of methods targets to update the \textbf{cross-attention layers in LDM}, which determine how textual prompts influence the generated visual output. FMN~\cite{Zhang2023ForgetMeNotLT} fine-tunes cross-attention module to re-steer attention mechanisms to eliminate certain concepts while maintaining generative quality. 
 AC~\cite{Ablating_Concept} introduces an anchor-based fine-tuning strategy, aligning erased concepts with broader semantic categories to suppress their stylistic or object-based representations.

Another set of methods fine-tunes the \textbf{LDM backbone}, directly modifying the denoising process to eliminate specific concepts from the model’s latent representations. 
ESD~\cite{esd} fine-tunes the LDM to match the noise prediction of \( c_{\text{erase}} \) to that of \( c_{\text{target}} \), ensuring erased concepts remain irrecoverable. This optimization is guided by classifier-free guidance (Eq.~\eqref{eq:classifier-free-guidance}), which directs the model's learning signal. During fine-tuning, ESD modifies either cross-attention or non-attention modules to reinforce robustness against adversarial prompts.
SALUN~\cite{fan2024salun} applies saliency-guided erasing, selectively updating high-impact weights to maximize forgetting while minimizing unintended side effects. 
DT~\cite{Ni2023DegenerationTuningUS} conditions the model to generate structurally degraded outputs when prompted with erased concepts, effectively neutralizing their representation in the latent space.

Several approaches incorporate geometric constraints, continual learning, or robustness against personalization to enhance fine-tuning methods. Geom-Erasing~\cite{Liu2023ImplicitCR} removes implicit visual concepts, such as watermarks and hidden signals, by introducing geometric constraints that disrupt structured artifacts without degrading unrelated content. SA~\cite{Heng2023SelectiveAA} leverages continual learning techniques, employing regularization-based forgetting to erase targeted concepts while preserving generalization and mitigating catastrophic forgetting. Lastly, IMMA~\cite{Zheng2023IMMAIT} adopts a preventive fine-tuning approach, modifying model weights preemptively to resist unauthorized adaptation via fine-tuning techniques, thereby preventing the downstream personalization of diffusion models for unethical or restricted purposes. Additionally, SPM~\cite{Lyu2023OnedimensionalAT} introduces a lightweight, one-dimensional adapter that enables precise and transferable concept erasure across different diffusion models.

\paragraph{Fine-tuning CLIP.}
Fine-tuning latent diffusion model showed great success for concept erasure, plus their interpretability to understand erasing concepts. However, to apply these methods to the updated LDM or other structured LDM, they have to get changed or redesigned since these methods are designed only to specific LDM. As one of the strength of fine-tuning CLIP model for concept erase, the CLIP text encoder whose concepts get erased, $\mathcal{E}_{\text{txt}}$, can transfer to the other structure LDM as long as they still depends on CLIP model. To fine-tune the CLIP model, \cite{safe_clip} generates a dataset composed of quadruplets of safe and unsafe text-image pairs (ViSU dataset). 
Similarly, \cite{Liu2024LatentGA} generates a dataset composed of safe and unsafe text pairs (CoPro dataset), where unsafe prompts are synthesized using a large language model, and safe counterparts are created by removing harmful concepts while preserving context.
After generating datasets, this study finetunes CLIP based on designed loss inspired by contrastive loss. This finetuned CLIP text encoder, $\mathcal{E'_{\text{txt}}}$, leads $\mathcal{E'}_{\text{txt}}(y_{erase}) \approx \mathcal{E}_{\text{txt}}(y_{target})$. Even if $y_{erase}$ is given to the SD model, it will generate images that are not aligned with concept to erase, $SD'(y_{erase}) \approx x_{target}$, where $SD'(y) = \mathcal{D}(\Tilde{\Phi}_{\theta}(z_T, \mathcal{E'}_{\text{txt}}(y)))$.
These methods offer better adaptability than LDM fine-tuning approaches. However, this approach requires a carefully curated and extensive dataset to fine-tune the existing text encoder, unlike other concept erasure methods.

\subsection{Closed-form Model Editing Methods} \label{subsec:closed}
Fine-tuning methods are intuitive and effective for modifying SD. However, they require iterative optimization through gradient descent, making them computationally expensive and time-consuming. Moreover, fine-tuning introduces risks of overfitting and unintended degradation of the model’s capabilities, necessitating careful hyperparameter tuning. In contrast, closed-form solutions provide a direct mathematical update to model parameters without iterative training. This enables faster application of model modifications while eliminating the need for extensive hyperparameter tuning. 

A general formulation of closed-form model editing follows a least squares based optimization:
\begin{equation} \label{eq:closed_form}
\min_{W} \left\| W c_{\text{erase}} - W_0 c_{\text{target}} \right\|_2^2,
\end{equation}
where \( W \) denotes the editable parameters of the model, primarily the key and value projection matrices in the cross-attention module, while \( W_0 \) represents the pre-trained weights. Closed-form solutions directly compute the optimal update for \( W \) directly, enabling efficient modification while preserving overall model coherence. To enhance stability and prevent unintended interference, regularization terms are incorporated into Eq.~\eqref{eq:closed_form}, balancing alignment with the target concept while minimizing deviations from the original model.

Notable closed-form methods include ReFACT~\cite{Arad2023ReFACTUT}, which applies a low-rank memory update to the CLIP text encoder, ensuring persistent factual knowledge updates while minimizing interference by unrelated concepts. TIME~\cite{Orgad2023EditingIA} modifies the LDM’s cross-attention projection matrices, aligning implicit assumptions in generated images with desired attributes. Unified Concept Editing (UCE)~\cite{Gandikota2023UnifiedCE} introduces a closed-form method for simultaneous multi-concept editing in T2I models, enabling scalable erasure, moderation, and debiasing by modifying cross-attention projections while minimizing interference against unedited concepts. MACE~\cite{Lu2024MACEMC} further refines cross-attention weights by integrating adapter-based concept erasure, achieving precise removal of up to 100 concepts in a more memory-efficient manner.

Other recent works have explored additional extensions of closed-form model editing. EMCID~\cite{Xiong2024EditingMC} introduces a two-stage framework combining self-distillation and closed-form updates, scaling to over 1,000 concurrent modifications. MUNBa~\cite{Wu2024MUNBaMU} formulates concept erasure as a Nash bargaining problem, deriving an equilibrium update that balances forgetting and preservation objectives.

Overall, closed-form methods offer a computationally efficient alternative to fine-tuning by providing direct parameter updates. These methods ensure fast and stable modifications.



\subsection{Inference-time Intervention Methods} \label{subsec:inference}
Both fine-tuning and closed-form methods demonstrate intuitive and effective performance in concept erasure. Fine-tuning LDM and closed-form parameter updates enable direct control over the generative process.
Among them,  CLIP fine-tuning provides a flexible plug-and-play concept erasure solution for T2I models that rely on CLIP encoders.

However, these approaches require weight modifications to SD components, limiting their adaptability and deployment efficiency. In contrast, inference-stage control methods enable concept erasure without modifying SD’s parameters. These methods instead intervene at the inference stage by modifying classifier-free guidance (Eq.~\eqref{eq:classifier-free-guidance}), editing textual embeddings \( c \), or sanitizing input prompts \( y \) using large language models.

\paragraph{Modifying Classifier-Free Guidance.}  
A core approach for inference-stage concept erasure is adjusting classifier-free guidance (Eq.~\eqref{eq:classifier-free-guidance}) to steer the generative process away from undesired content. Safe Latent Diffusion~\cite{sld} modifies the classifier-free guidance signal in SD’s denoising process, redirecting latent activations to prevent the generation of unsafe concepts. Anti-Memorization Guidance~\cite{Chen2024TowardsMD} introduces despecification and dissimilarity constraints that adjust classifier-free guidance dynamically, ensuring that models do not overfit to specific training instances or regenerate memorized images. Dynamic Negative Guidance~\cite{koulischer2024dynamic} further refines this strategy by adaptively estimating negative guidance scales at each sampling step, enabling fine-grained suppression of target concepts without additional training. Training-Free Safe Denoising (TFSD)~\cite{Kim2025TrainingFreeSD} proposes an alternative formulation of the denoising function itself, incorporating image-based unsafe priors to penalize sampling trajectories near unsafe regions. All these methods leverage guidance re-weighting strategies to suppress undesired features while maintaining high image quality.

\paragraph{Editing Textual Embeddings.}  
Rather than modifying the diffusion process, another class of inference-stage methods operates on text embeddings to enforce concept erasure. SAFREE~\cite{safree} applies subspace projection and adaptive re-attention to detect and suppress undesirable content within CLIP text embeddings before they get used for image synthesis. Similarly, Content Suppression in T2I Models~\cite{Li2024GetWY} employs soft-weighted regularization to refine textual embeddings during sampling, ensuring that forbidden concepts do not appear in generated outputs. These methods enable fine-grained, token-level control over the generation process while preserving overall model flexibility.

\paragraph{Sanitizing Input Prompts Using LLMs.}  
A third category of inference-stage control leverages Large Language Models (LLM) to preprocess prompts, ensuring that user inputs do not contain prohibited content before the diffusion process begins. ORES ~\cite{ores} employs LLM-based query rewriting to automatically sanitize user prompts, replacing restricted terms with conceptually aligned yet safe alternatives. On the other hand, GuardT2I~\cite{Yang2024GuardT2IDT} detects adversarial prompts that attempt to bypass safety mechanisms, leveraging a fine-tuned LLM to analyze and reject unsafe queries before image generation. 

By operating externally to the SD components and modifying only the diffusion process at inference, these methods remain model-agnostic and scalable across different T2I model architectures.

\section{Robustness} \label{sec:robustness}
Concept erasure methods aim to prevent T2I models from generating undesired concepts. For instance, when a model undergoes fine-tuning to erase a copyrighted artist’s style (e.g., Van Gogh), prompts such as ``cypresses by Van Gog'' should ideally produce outputs that bear no resemblance to the artist’s original style, as shown in Fig.~\ref{fig:overview}.

However, studies demonstrate that minor perturbations to the prompt—such as the addition of unrelated tokens or imperceptible modifications—can effectively circumvent concept erasure, allowing target T2I model to regenerate removed concepts. In some cases, even semantically meaningless inputs would  exploit the underlying representations within SD models to reconstruct erased concepts.

We categorize adversarial attacks based on whether they require access to the Latent Diffusion Model (LDM) of Stable Diffusion (SD). Following this, we also discuss existing defense strategies against such attacks.

\subsection{Adversarial Attacks} \label{subsec:adv_atk}
Adversarial attacks against T2I models manipulate prompt inputs or textual embeddings to reconstruct erased concepts. Given a concept-erased model, an adversarial prompt is optimized to elicit outputs that closely resemble those generated by an unaltered model. A general formulation of adversarial attacks against concept erasure is:
\begin{equation}
\underset{z_{\text{adv.}} \text{ or } y_{\text{adv.}}}{\mathrm{argmin}}  || \mathcal{SD'}(z_{\text{adv.}},y_{\text{adv.}}) - x_{\text{erase}} ||,    
\end{equation}
where \( \mathcal{SD'} \) denotes the concept-erased Stable Diffusion model, and \( z_{\text{adv.}} \) and \( y_{\text{adv.}} \) represent adversarial latents and prompts, respectively, that restore the erased concept within \( \mathcal{SD'} \).

\paragraph{Attacks with LDM Access}
These attacks access LDM’s latent representations, enabling adversaries to design strategies that systematically bypass concept erasure mechanisms. While such access is rare in real-world scenarios, these attacks remain essential for stress-testing erasure techniques.

One class of such attacks invert concept erasure transformations to recover removed concepts.  
Circumventing Concept Erasure~\cite{pham2024circumventing} optimizes adversarial embeddings via inversion using LDM, successfully retrieving erased concepts. Similarly, Concept Arithmetics~\cite{Petsiuk2024ConceptAF} reconstructs erased concepts by manipulating latent representations and leveraging semantic composition to synthesize forbidden attributes through linear combinations of concept embeddings.

Another category of attacks exploits prompt tuning and adversarial optimization to bypass safety-driven concept removal. P4D~\cite{p4d} systematically tunes adversarial prompts by iteratively refining textual inputs based on model feedback, demonstrating that safety filters and erasure techniques remain vulnerable to adversarial prompt engineering. Additionally, UnlearnDiffAtk~\cite{to_generate_or_not} specifically targets models trained with safety-driven unlearning by crafting prompts that reverse-engineer unlearning constraints, showing that adversarially optimized inputs can still induce the generation of prohibited content.

Although these attacks require privileged access to LDM, they provide valuable insights into the transferability of adversarial prompts across models. By assessing how an attack generalizes to different versions of SD and CLIP, these studies reveal broader vulnerabilities in concept erasure methods.

\paragraph{Attacks without LDM Access}  
Even without LDM access, adversarial methods effectively bypass concept erasures by manipulating prompt or textual embeddings, without interacting with the diffusion model’s internal denoising process.

PEZ~\cite{hard_prompt} formulates a discrete optimization problem to recover a text prompt that closely aligns with a given erased concept image, \( x_{\text{erase}} \), leveraging CLIP’s vision-language similarity for optimization without requiring access to the underlying diffusion model. 
Similarly, MMA-Diffusion~\cite{Yang2023MMADiffusionMA}, like PEZ, employs a surrogate model for adversarial attacks, operating within CLIP’s text encoder embedding space or its multi-modal space. By optimizing adversarial prompts and introducing imperceptible perturbations, MMA-Diffusion successfully bypasses safety mechanisms, demonstrating its effectiveness in both text-to-image generation and image editing tasks.

Ring-A-Bell~\cite{Tsai2023RingABellHR} proposes a model-agnostic red-teaming framework that reconstructs erased concepts by optimizing adversarial prompts using a genetic algorithm in the text embedding space. Likewise, RIATIG~\cite{Liu2023RIATIGRA} employs a genetic optimization strategy to iteratively refine adversarial queries, enabling content moderation evasion across multiple T2I models. 
Meanwhile, UPAM~\cite{Peng2024UPAMUP} utilizes gradient-based prompt tuning combined with semantic-enhancing learning to systematically generate adversarial prompts capable of bypassing API-level safety mechanisms.

These approaches expose a fundamental vulnerability in concept erasure techniques—even without LDM access, adversarial optimization in the prompt space alone can effectively reconstruct erased content. This underscores the need for stronger prompt filtering mechanisms and adversarially resilient diffusion models to prevent circumvention through external manipulations.

\subsection{Defensive Methods}
To strengthen concept-erased models against adversarial prompt attacks (Sec.\ref{subsec:adv_atk}), recent research integrates adversarial training into concept erasure techniques, enhancing robustness while preserving generation fidelity. These defenses align with the categorization of concept erasure methods (Fig.\ref{fig:taxonomy}, Tab.\ref{tab:taxonomy}), targeting distinct architectural components. 
By addressing vulnerabilities across these optimization spaces, defensive methods improve the reliability of concept erasure while maintaining the expressiveness of T2I models.

R.A.C.E.\cite{RACE} employs a single-timestep adversarial attack to efficiently identify vulnerabilities in SD and leverages this attack for adversarial fine-tuning, significantly reducing attack success rates in both white-box and black-box settings. Receler\cite{Huang2023RecelerRC} integrates a lightweight robust eraser within cross-attention layers of LDM, utilizing concept-localized regularization and adversarial prompt learning to improve robustness against paraphrased attacks while preserving non-target concepts. AdvUnlearn~\cite{Zhang2024DefensiveUW} advances the robust erasing paradigm by applying adversarial training on the CLIP text encoder, enhancing prompt-space robustness while maintaining the alignment between textual prompts and image generation. RECE~\cite{Gong2024ReliableAE} extends closed-form concept erasure by incorporating adversarial fine-tuning on matrix-modified cross-attention layers, efficiently discovering and erasing adversarial embeddings in a fully closed-form manner. Additionally, in inference-stage control, SAFREE~\cite{safree} demonstrates strong robustness compared to other defense methods, despite not employing adversarial training.

By integrating adversarial robustness into concept erasure, these methods significantly improve the reliability of T2I safety mechanisms. However, challenges remain in optimizing the balance between robustness, generation quality, and computational efficiency, warranting further exploration in adaptive adversarial training strategies for future diffusion models.

\section{Evaluation} \label{sec:evaluation}
Evaluating concept erasure methods is essential for quantifying their effectiveness and enabling fair comparisons across different approaches. This section reviews widely adopted metrics and datasets to assess both the success of concept removal and the preservation of general model capabilities.

\subsection{Metrics}
Concept erasure methods are typically evaluated in two key aspects: erasure effectiveness and model fidelity. 

The Erasure Success Rate (ESR) measures how effectively a method removes a target concept. This is commonly assessed using classification accuracy, where a pre-trained classifier determines whether the erased concept remains present in the generated images. Formally, ESR is defined as:
\begin{equation}\label{eq:esr}
    \text{ESR} = \frac{1}{N} \sum_{i=1}^{N} \mathbbm{1}\left( f\left(SD'(y_i)\right) = c_{\text{erase}} \right),
\end{equation}
where \( y_i \) represents the prompt, \( c_{\text{erase}} \) is the erased concept, \( f \) is a classifier, and \( N \) is the total number of prompts. Lower ESR values indicate more successful erasure. ESR can also be extended to evaluate robustness against adversarial attacks by replacing standard prompts \( y_i \) with adversarially optimized prompts \( y_{\text{adv.}} \) or modifying latent variables \( z_{\text{adv.}} \), allowing an assessment of how well the model resists attempts to reconstruct erased concepts.

To ensure that erasure does not degrade the model's ability to generate non-erased content, model fidelity is evaluated by measuring both image quality and text-image alignment before and after concept removal. Fréchet Inception Distance (FID)~\cite{fid} is widely used to quantify changes in the perceptual quality of generated images. In addition to image quality, maintaining alignment between textual prompts and generated outputs is crucial. CLIP score~\cite{hessel2021clipscore} is commonly employed for this purpose, providing a similarity measure between the generated image and its corresponding textual prompt. Furthermore, ESR can also serve as a fidelity metric by computing it on prompts unrelated to the erased concept, denoted as \( y_{\text{non-erase}} \).

\subsection{Datasets}
Dataset selection depends on the nature of the concepts being erased, with commonly used datasets categorized according to their evaluation objectives. For assessing NSFW content removal, the I2P dataset~\cite{Schramowski2022SafeLD}, which consists of 4,703 real-world user-generated prompts, is widely employed. Object concept removal is typically evaluated using structured prompts such as ``A photo of [object class]'', enabling controlled experiments on whether erased objects still appear in generated images. Artistic style erasure often relies on ESD's artist prompt dataset~\cite{esd}, which provides standardized prompts referencing specific artistic styles.

To evaluate model fidelity, the COCO dataset~\cite{lin2014microsoft} is commonly used. This dataset enables FID-based image quality assessment and supports CLIP score evaluation for measuring text-image alignment. Beyond standard datasets, robustness evaluation requires datasets explicitly designed for testing adversarial vulnerabilities. 
For example, the MMA-Diffusion benchmark~\cite{Yang2023MMADiffusionMA} and Ring-A-Bell dataset~\cite{Tsai2023RingABellHR} feature adversarial prompts designed to evade concept erasure and systematically test its vulnerabilities.

Together, these metrics and datasets establish a comprehensive framework for evaluating concept erasure, ensuring that methods are assessed not only for their effectiveness in removing targeted concepts but also for their ability to maintain generative quality and resist adversarial attacks.


\section{Future Research Directions} \label{sec:future_research}

Concept erasure techniques for T2I models have demonstrated promising results in mitigating the generation of undesired content. However, several open challenges remain. In this section, we outline key future research directions that can further advance the development of responsible and robust generative models.

\paragraph{Extending to Other Modalities:} While this survey primarily focuses on concept erasure for T2I models, extending these techniques to other generative modalities—such as text-to-video and text-to-audio—remains an open challenge. Despite structural similarities, these modalities introduce additional complexities, such as temporal consistency in videos and waveform coherence in audio. Understanding how concept erasure methods generalize across different modalities and identifying modality-specific adaptations will be crucial for developing responsible generative AI systems.

\paragraph{Designing More Comprehensive Benchmarks:} Existing evaluations primarily assess erasure effectiveness by detecting the presence of erased concepts in generated images. However, the broader impact of concept erasure on non-target concepts remains under-studied. For instance, erasing the concept of ``Van Gogh'' may inadvertently alter representations of other Impressionist artists. Future benchmarks should move beyond dataset-level assessments and introduce fine-grained, concept-level evaluations that capture unintended effects on related visual attributes. Establishing standardized, interpretable, and reproducible benchmarks will enable more rigorous comparisons across different erasure techniques.

\paragraph{Enhancing Robustness Against Adversarial Attacks:} Current concept erasure methods remain vulnerable to adversarial attacks that exploit weaknesses. While various defensive strategies have been proposed, adversarial attacks continue to evolve, exposing new vulnerabilities. Future work should focus on developing more robust and adaptive defenses that generalize across different attack strategies. This may include integrating adversarial training, certifiable robustness techniques, and multi-modal alignment to mitigate adversarial circumvention while preserving model fidelity.

By addressing these challenges, future research can contribute to the development of more responsible, robust, and generalizable concept erasure techniques for T2I models and beyond.

\section{Conclusion} \label{sec:conclusion}
Concept erasure in Text-to-Image (T2I) models is crucial for ensuring ethical and legal compliance in generative AI. This survey categorizes existing approaches based on their optimization strategies and modified model components, covering fine-tuning, closed-form solutions, and inference-time interventions. We also discuss adversarial attacks and emerging defenses. Despite progress, challenges remain in balancing erasure effectiveness, model fidelity, and adversarial robustness. Expanding concept erasure to other modalities, refining evaluation benchmarks, and developing stronger defenses are key directions for future research. We hope this survey serves as a foundation for advancing responsible and secure generative AI.

\bibliographystyle{named}
\bibliography{refs}

\end{document}